\documentclass{article}
\usepackage{spconf,amsmath,epsfig}
\usepackage{graphicx}
\usepackage{subfigure}
\usepackage{multirow}
\usepackage{color}
\usepackage{changepage}
\usepackage{times}
\usepackage{float}
\usepackage{amssymb}
\usepackage{multirow}
\usepackage{bbding}
\usepackage{booktabs}
\usepackage{caption}
\usepackage{cite}
\usepackage{array}
\usepackage{adjustbox}
\usepackage[section]{placeins}
\usepackage{hyperref}

\title{TransClaw U-Net: Claw U-Net with transformers for medical image segmentation}
\name{Chang Yao$^1$, Menghan Hu$^{1,*}$, Guangtao Zhai$^2$, Xiao-Ping Zhang$^3$}

\address{$^1$Shanghai Key Laboratory of Multidimensional Information Processing, East China Normal University \\
$^2$Institute of Image Communication and Information Processing, Shanghai Jiao Tong University\\
$^3$Department of Electrical, Computer and Biomedical Engineering, Ryerson University\\
\thanks{This work is sponsored by the Shanghai Education Development Foundation and Shanghai Municipal Education Commission (No. 19CG27).}
\thanks{The authors thank the developers of TransUNet for their technical support.}
\thanks{Corresponding author: Menghan Hu (mhhu@ce.ecnu.edu.cn)}
}

\begin{document}
%\ninept
%

\maketitle
\begin{abstract}
In recent years, computer-aided diagnosis has become an increasingly popular topic. Methods based on convolutional neural networks have achieved good performance in medical image segmentation and classification. Due to the limitations of the convolution operation, the long-term spatial features are often not accurately obtained. Hence, we propose a TransClaw U-Net network structure, which combines the convolution operation with the transformer operation in the encoding part. The convolution part is applied for extracting the shallow spatial features to facilitate the recovery of the image resolution after upsampling. The transformer part is used to encode the patches, and the self-attention mechanism is used to obtain global information between sequences. The decoding part retains the bottom upsampling structure for better detail segmentation performance. The experimental results on Synapse Multi-organ Segmentation Datasets show that the performance of TransClaw U-Net is better than other network structures. The ablation experiments also prove the generalization performance of TransClaw U-Net. 

\end{abstract}
\section{Introduction}         
\label{sec:intro}
With the continuous development of deep learning, computer vision has been widely applied in many fields, including manufacturing, automatic driving, computer-aided diagnosis and agriculture. Specially, computer aided diagnosis has attracted the attention of many scholars at home and abroad. Many methods based on convolutional neural network (CNN) have been used in image classification and segmentation \cite{long2015fully}, and have achieved good performance. In particular, the accurate medical image segmentation plays an important role in clinical diagnosis \cite{hatamizadeh2021unetr}. Among various network structures, U-Shape based UNet \cite{ronneberger2015u} framework has achieved great success in the field of medical image segmentation. The encoder of UNet provides a high-level semantic feature map, and the decoder provides a low-level detailed feature map. These two phases are combined through skip connections \cite{yao2020claw}. Many improved network structures based on U-Net have achieved better results in different medical datasets, such as UNet++ \cite{zhou2019unet++}, Resnet34-UNet \cite{xiao2018weighted}, Channel-UNet \cite{chen2019channel}, Attention-UNet \cite{oktay2018attention}, and R2UNet \cite{alom2018recurrent}, which can fully learn the rich detail information in the image and restore it based on the location information. The success of these studies in organ segmentation \cite{li2018h}, lesion segmentation \cite{zhou2017fixed} and cardiac segmentation \cite{yu2017automatic} have fully proved the high efficiency and superiority of CNNs.

Although most of the current medical image segmentation tasks are based on CNNs, it is still difficult to meet the requirements of accurate segmentation in computer-aided diagnosis. The main reason is that CNN makes full use of the end-to-end learning method, and it lacks the modeling of some long-range ordered features in the images \cite{chen2021transunet}, which leads to the poor segmentation performance of many regions. Each convolution kernel only pays attention to the feature information of itself and its boundary, and lacks a large range of feature fusion, thus affecting the overall effect. To overcome these disadvantages of convolution operation, many scholars put forward some novel solutions, including the self-attention mechanism \cite{schlemper2019attention}, atrous convolution \cite{gu2019net} and image pyramid \cite{zhao2017pyramid}.These methods does not consider the long-range dependencies of medical images.

In natural language processing (NLP), the prevalent architecture transformer demonstrates strong learning capabilities for long-term features such as BERT \cite{devlin2018bert} and GPT \cite{radford2018improving} models in sequence modeling and transduction tasks \cite{liu2021swin}. For the transformer operation, the distributed convolution operator is combined with the multi-head attention mechanism, which can find the relationship among different input parts \cite{vaswani2017attention}. Many researches have started to use the transformer structure in computer vision. Dosovitskiy et al \cite{dosovitskiy2020image} proposed visual transformer (ViT) technology for image recognition tasks. Liu et al \cite{liu2021swin} proposed a hierarchical swin transformer structure, which uses the shift window for feature calculation and achieves better results in image classification. Recently, transformer has also been used in medical image segmentation tasks and has achieved satisfactory results. Chen et al \cite{chen2021transunet} inserted transformer into the encoding layer of UNet structure, which encodes patches based on feature maps so as to realize the characteristics of long-term dependency. Cao et al \cite{cao2021swin} proposed a UNet-like pure transformer architecture, which uses transformer with drifted windows for feature extraction and patch-expanding transformer for restoration of image resolution.

In this research, we propose the TransClaw U-Net architecture, which combines the detailed segmentation ability of Claw U-Net with the long-term feature learning ability of transformer structure, and apply it to medical image segmentation. This network uses the convolution operation to obtain the feature map, and then encodes the patches of high-resolution images using the feature map. Transformer strengthens the global connection between the encoding parts, which makes the image features sufficiently visible, and the self-attention mechanism is more conducive to the detail segmentation. In the decoding part, the bottom upsampling module of Claw U-Net is retained, and the deep feature map is combined with the shallow layer to achieve accurate positioning. Finally, the encoding part, the upsampling part and the decoding part of the corresponding layers are combined to achieve the restoration of image resolution. Experimental results on abdominal CT scans datasets show that our network structure achieves more accurate organ segmentation. Meanwhile, we have also verified that larger input image size, smaller patch size and more skip-connections can improve the model performance, but they require more computing resources.

\section{RELATED WORKS}
CNN and variants: As the basic framework of deep learning, CNN has gradually replaced the traditional machine learning methods \cite{tsai2003shape} and become the most effective methods in the field of medical image segmentation due to its powerful image learning ability. Since 2015, UNet structure has been proposed and proved to have superior performance in medical image segmentation tasks. More and more new network structures based on U-shape have been proposed, which include Resnet34-UNet \cite{xiao2018weighted}, Attention-UNet \cite{chen2019channel}, Dense-UNet \cite{li2018h}, and UNet++ \cite{zhou2019unet++}. Some improved convolution operations are introduced to improve the model performance, such as deformable convolution \cite{xie2017aggregated} and deep convolution \cite{zhu2019deformable}. CNN and its variants have become the most primary back bone architecture in the field of artificial intelligence due to their strong learning ability.

Self-attention/Transformers to complement CNNs: In recent years, there are a lot of researches on how to integrate self-attention mechanisms or transformer operations with CNNs. By encoding the images, the self-attention mechanism can make up for the deficiency of CNNs' long-range dependence and heterogeneous interactions, thus improving the head network \cite{hu2018relation} and back bone network \cite{wang2018non}. Wang et al \cite{wang2018non} inserted new non-local operators into the non-local operators of multiple intermediate convolution layers. By adding transformer structure into UNet structure and introducing global self-attention mechanism, Chen et al \cite{chen2021transunet} have achieved good results in 2D CT image dataset. Based on the above researches, we adopt the newly proposed Claw U-Net structure \cite{yao2020claw} as the framework, and introduce the transformer structure to explore its effect in medical image segmentation task.

Transformers: the transformer structure was first used in the NLP domain, and has achieved excellent results in machine translation \cite{vaswani2017attention}. As it has strong ability to model the long-term features of sequences, it can make up for some deficiencies of CNNs in global features. Many scholars have started to combine transformer structures with convolution operations to achieve better performance. Recently, in imagenet classification tasks, ViT \cite{dosovitskiy2020image} has achieved a great break-through by applying the global attention mechanism to images. However, the network results of the UNet structure can be trained on small datasets, the transformer structure requires a large amount of data to be pre-trained. Deit \cite{touvron2020training} proposed several training strategies for applying ViT to small datasets. In the structure proposed in this paper, the Claw U-Net network structure is adopted as the skeleton, and the convolution operation is first carried out in the encoding part. Transformer structure is introduced to the obtained feature maps to encode them, and the corresponding features are extracted through the multi-head attention mechanism. The decoding part combines the bottom upsampling feature maps with the encoding feature maps and the decoding feature maps to restore the spatial features of the whole image.

\section{Method}
In this section, we will introduce the overall structure of the TransClaw U-Net. This network introduces the transformer structure in the encoding layer, and uses the self-attention mechanism to achieve better global feature learning. During pixel restoration, it combines encoding, bottom up-sampling, and decoding part, so that detailed information can be maintained, and finally the image segmentation task is realized.

\subsection{Overall structure of TransClaw U-Net}

\begin{figure*}[htbp] \centering 

\includegraphics[height=8cm, width=18cm]{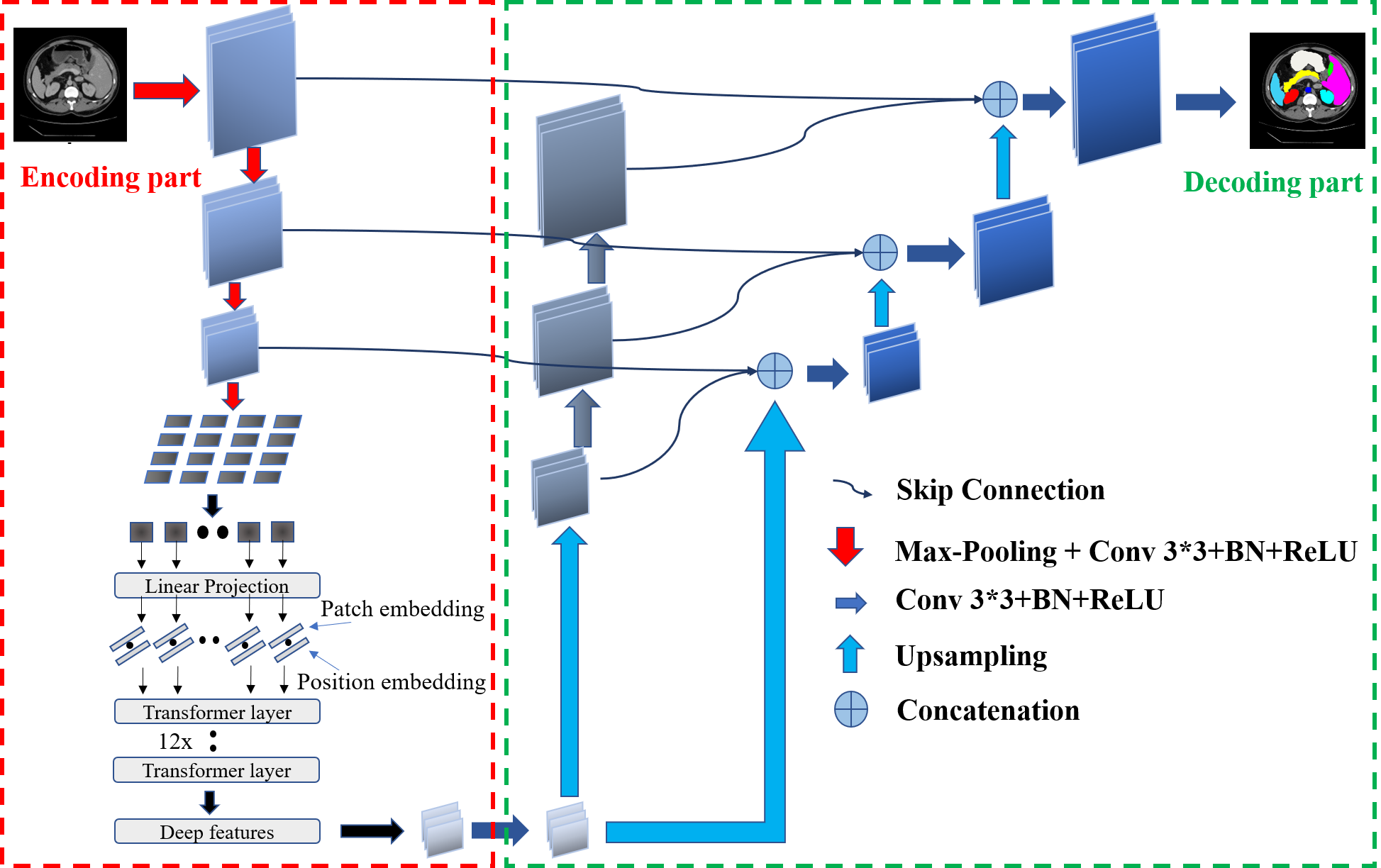} 

\caption{The architecture of TransClaw U-Net} \label{fig:graph} \end{figure*}

The overall structure of TransClaw U-Net is shown in Fig. 1, which mainly includes the encoding part, the bottom upsampling part and the decoding part. These parts are connected by skip-connections.  Compared with Claw U-Net, TransClaw U-Net uses a combination of convolution and transformer mechanism in the encoding part. In the encoding part, assuming that $I_{in}$ denotes the input image, $I_{fea}$ denotes the feature images after convoluton operation, $I_{deep}$ denotes the deepest image in the encoding part. For the convolution parts, the operation can be formulated as:

\begin{equation}
\mathrm{F}_{0}=\sigma\left(B N\left(\operatorname{Conv}\left(\mathrm{I}_{i n}\right)\right)\right)
\end{equation}

where $\sigma$ represents the ReLU function, $BN$ represents the batchnormalization function, and $conv$ represents thr convolution function. Before each convolution parts, maxpooling operation is used to reduce the feature map size by two times.

\begin{equation}
\mathrm{F}_{1}=\operatorname{MaxPool}\left(\mathrm{I}_{\text {fea }}\right)
\end{equation}

\begin{table*}[htbp]
\centering 
\caption{\label{tab:test}Comparison of model performance on the Synapse multi-organ CT dataset. The best performers are highlighted in bold.}
 \scalebox{1}[1]{
\resizebox{\textwidth}{!}{
 \begin{tabular}{lccccccccccc} 
  \toprule 
  Model & DSC & HD & Aorta & Gallbladder & Kidney(L) & Kidney(R) & Liver & Pancreas & Spleen & Stomach  \\ 
  \midrule 
 V-Net \cite{milletari2016v}  & 68.81 & - & 75.34 & 51.87 & 77.10 & \textbf{80.75} & 87.84 & 40.05 & 80.56 & 56.98  \\ 
 DARR \cite{fu2020domain} & 69.77 & - & 74.74 & 53.77 & 72.31 & 73.24 & 94.08 & 54.18 & \textbf{89.90} & 45.96 \\ 
 U-Net \cite{ronneberger2015u} & 76.85 & 39.70 & \textbf{89.07} & \textbf{69.72} & 77.77 & 68.60 & 93.43 & 53.98 & 86.67 & 75.58  \\ 
 R50 U-Net \cite{chen2021transunet} & 74.68 & 36.87 & 87.74 & 63.66 & 80.60 & 78.19 & 93.74 & 56.90 & 85.87 & 74.16 \\  
 R50 Att-UNet \cite{chen2021transunet} & 75.57 & 36.97 & 55.92 & 63.91 & 79.20 & 72.71 & 93.56 & 49.37 & 87.19 & 74.95 \\ 
 R50 ViT \cite{chen2021transunet} & 71.29 & 32.87 & 73.73 & 55.13 & 75.80 & 72.20 & 91.51 & 45.99 & 81.99 & 73.95 \\ 
 TransUNet \cite{chen2021transunet} & 77.48 & 31.69 & 87.23 & 63.13 & 81.87 & 77.02 & 94.08 & 55.86 & 85/08 & \textbf{75.62}  \\ 
 TransClaw U-Net & \textbf{78.09} & \textbf{26.38} & 85.87 & 61.38 & \textbf{84.83} & 79.36 & \textbf{94.28} & \textbf{57.65} & 87.74 & 73.55  \\ 
 
  \bottomrule 
 \end{tabular}}}
 
\end{table*}

where $MaxPool$ represents the maxpooling function. The purpose of these three convolution operations is to retain low-level feature maps, so that the shallow features and contours in the medical image can be effectively combined with the decoding part. To make TransClaw U-Net learn better long-term spatial features, we introduce a transformer mechanism after the convolution parts which can be summarized as:

\begin{equation}
\mathrm{I}_{\text {deep }}=\operatorname{Trans}\left(\mathrm{I}_{\text {fea }}\right)
\end{equation}
 
where $Trans$ represents the transformer operation, which will be described in detail in section 3.2. After the transformer module, the size of the encoded image is restored to $H/P\times W/P$ (where $H$,and $W$ is the height and width of the input image, and $P$ is the patch size) and sent to the decoding part and the bottom upsampling part.

In the decoding part, the feature maps can be restored to the original resolution after several upsampling operations. The structure of TransClaw U-Net combines the corresponding encoding part, the bottom upsampling part, and the decoding part through skip-connections. The expression to skip the link can be expressed as:

when $i=1, \cdots, N-1$,
\begin{small}
\begin{equation}
\mathrm{x}_{\mathrm{De}}^{\mathrm{i}, 2}=\left[\mathrm{C}\left(\mathcal{D}\left(X_{\mathrm{En}}^{\mathrm{k}, 0}\right)\right)_{k=1}^{i-1}, \mathrm{C}\left(X_{\mathrm{Up}}^{\mathrm{i}, 1}\right), \mathrm{C}\left(U\left(X_{\mathrm{De}}^{\mathrm{k}, 2}\right)\right)_{k=i+1}^{N}\right]
\end{equation}
\end{small}

when $i = N$,
\begin{small}
\begin{equation}
\mathrm{x}_{\mathrm{De}}^{\mathrm{i}, 2}=\mathrm{x}_{\mathrm{En}}^{\mathrm{i}, 0}
\end{equation}
\end{small}

where $i$ represents the downsampling layer along with the encoder, $N$ represents the total layer of the encoder, $x_{De}^{i,2}$ represents the decoding part of the $i-th$ layer, $x_{En}^{i,0}$ represents the encoding part of the $i-th$ layer, $x_{Up}^{i,1}$ represents the decoding part of the $i-th$ layer. Let function $C(\cdot)$  denotes a convolution operation,  $\mathcal{D}(\cdot)$ and $U(\cdot)$ denote upsampling and downsampling operation separately,  $[\cdot]$ denotes connecting together.

Each upsampling operation will go through a $3\times3$ convolution operation before. After the image is restored to $H/2 \times W/2$, no concatenation operation is required for the last upsampling. After that, the image size goes to the initial size.

\subsection{Transformer in TransClaw U-Net}

To preserve the initial features of the image and the contours of the organs, several convolutions are performed before the transformer operation, and the feature maps are used as input of transformer. Divide the image into $N$ patches of $P \times P$ size, which have no overlapping with each other, forming a flatten 2D sequence$\left\{\mathbf{x}_{p}^{i} \in \mathbb{R}^{P^{2} \cdot C} \mid i=1, \cdots, N-1\right\}.$

Concat the vectors of $N$ patches that are reshaped together, and we can get the $N \times\ P^{2}$ matrix, which is similar to the word vector input to the transformer. $N$ is the number of patches, which determines the length of the input sequence. It is worth noting that when the size of the patch changes, the length of the $P^{2}$ dimensional vector will also change after each patch is reshaped e. 

The structure in transformer is similar to that in NLP, which consists of $L-th$ layer transformer blocks. For the first layer of the encoder, assuming that the input is $z_{l-1}$ and the output is $z_{l}$, the results after they have passed through the Multihead Self-Attention (MSA) and Multi-Layer Perceptron (MLP) blocks can be represented by formulas 3 and 4, respectively.

\begin{equation}
\begin{aligned}
\mathbf{z}_{\ell}^{\prime} &=\operatorname{MSA}\left(\mathrm{LN}\left(\mathbf{z}_{\ell-1}\right)\right)+\mathbf{z}_{\ell-1}, & \ell &=1 \ldots L \\
\end{aligned}
\end{equation}

\begin{equation}
\begin{aligned}
\mathbf{z}_{\ell} &=\operatorname{MLP}\left(\operatorname{LN}\left(\mathbf{z}_{\ell}^{\prime}\right)\right)+\mathbf{z}_{\ell}^{\prime}, & \ell &=1 \ldots L
\end{aligned}
\end{equation}

where $L(\cdot)$ represents the layer normalization, and $z_{l}$ represents the images after encoding.

The Multi-Head Attention used in the transformer block is the core of the encoding. The study found that it is better to train multiple value vectors in parallel and then stitch them together to get the output than to use an attention function to get the $d_{model}$ dimensional value vector. Specifically, given an input matrix, we calculate multiple sets of $V$, $K$, and $Q$ matrices based on different parameter matrices, and then calculate multiple weighted $V$ matrices through multiple attention functions, and finally concatenate these matrices to get the final output through a weight matrix $W^{O}$. The above process can be expressed by the formula as follows,

\begin{equation}
\text { MultiHead }(Q, K, V)=\text { Concat }\left(\text { head }_{1}, \ldots, \text { head }_{\mathrm{b}}\right) W^{O}
\end{equation}

\begin{equation}
\text { head }_{\mathrm{i}}=\text { Attention }\left(Q W_{i}^{Q}, K W_{i}^{K}, V W_{i}^{V}\right)
\end{equation}

where $W_{i}^{Q} \in \mathbb{R}^{d_{\text {model }} \times d_{k}}, W_{i}^{K} \in \mathbb{R}^{d_{\text {model }} \times d_{k}}, W_{i}^{V} \in \mathbb{R}^{d_{\text {model }} \times d_{v}}, W^{O} \in \mathbb{R}^{h d_{v} \times d_{\text {model }}}$.

Since the dimensionality of the vector in each attention layer is equally divided, the computational cost of this method is not much different than that of using a single head attention.

\section{Experiments}
\subsection{Datasets}
Synapse Multi-organ Segmentation Dataset (Synapse): From the MICCAI 2015 Multi-Atlas Abdomen Labeling Challenge, 30 cases of abdominal CT scans were used, and there were a total of 3779 axial contrast-enhanced abdominal clinical CT images. 18 (2212 slices) of 30 cases are taken as the training dataset, and 12 (1567 slices) as the testing dataset. Following \cite{chen2021transunet} \cite{cao2021swin}, We used average Hausdor Distance (HD) and average Dice-Similarity coeffcient (DSC) as evaluation indicators to evaluate the segmentation performance of 8 organs in CT images including aorta, gallbladder, spleen, kidney(L), kidney(R), liver, pancreas, spleen and stomach.

\begin{figure*}[htbp] \centering 
	
	\includegraphics[height=10cm, width=18cm]{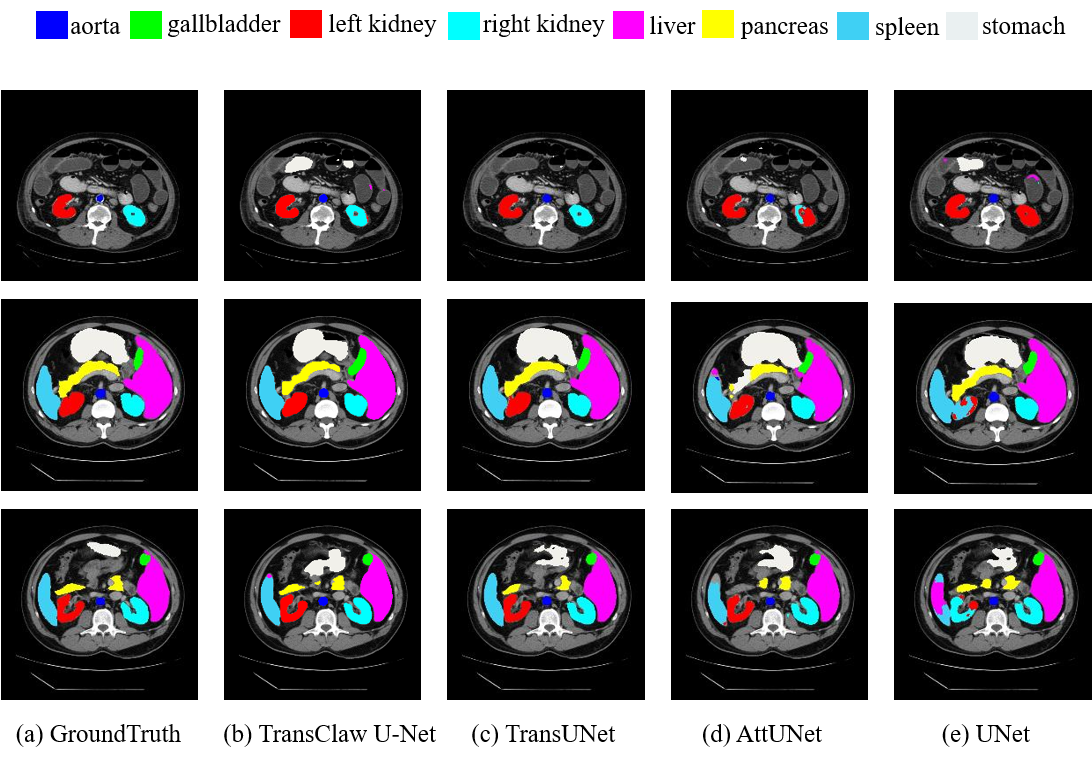} 
	
	\caption{Comparison of segmentation results of different models on the Synapse multi-organ CT dataset.} \label{fig:graph} \end{figure*}

\subsection{Implementation Details}
In this experiment, the proposed TransClaw U-Net is implemented in Python using the Pytorch deep learning framework and trained on the Synapse multi-organ CT dataset. In addition, the Stochastic Gradient Descent (SGD) is used with an initialized learning rate $1\times10^{-2}$ and momentum  $0.9$. The weight decay is set to $1\times10^{-4}$. Considering that the size of the input image is $224\times224$, we set the batch size to $24$, and train $150$ times. For a fair comparison, the parameters of all experiments are set to the same situation. In the training process, the model with the best performance on the validation set is selected as the final model. We use cross entropy as the loss function for optimization. All the experiments are performed on NVIDIA GeForce RTX 2080 Ti.

\subsection{Comparison with other models on Synapse dataset}

We test TransClaw U-Net on the Synapse multi-organ CT dataset and compare it with several state-of-the-art models, including: 1) V-Net \cite{milletari2016v}, 2) DARR \cite{fu2020domain}, 3) U- Net \cite{ronneberger2015u}, 4) AttnUNet \cite{oktay2018attention} and 5)TransU-Net\cite{chen2021transunet}. Table 1 shows the comparison results of various networks on the testing set. It can be clearly seen that our TransClaw U-Net has achieved the best results on the evaluation indicators with DSC and HD of $78.09\%$ and $26.38\%$, respectively. The DSC indicators of other networks are all lower than $77.5\%$ and the HD of other networks are all above $31.5$. Compared with the newly proposed TransUNet, our model improves the accuracy of DSC by $0.6\%$ and the accuracy of HD by $5.3\%$, which proves the effect of the upsampling part of the bottom layer on edge detail segmentation. It can also be seen that our model has the best segmentation performance in three of the 8 organs, while TransUNet only has one, which proves the effectiveness of our model.

Fig. 2 shows in detail the segmentation results of different network structures on the Synapse dataset. It can be seen that Attention-UNet has a problem of excessive segmentation of organs, and UNet's segmentation is not accurate. This is mainly due to the limitations of the convolution operation and the lack of the ability to extract global features of the image. Compared with TransUNet, our model has a better effect on the segmentation of organ boundaries, especially between organs that occupy a small area. Claw U-Net's preservation of image detail features and transformer's effective learning of global context information enable TransClaw U-Net to achieve better accurate medical image segmentation tasks.

\subsection{Ablation Experiments}
To discuss the impact of some input parameters used in the model on the results, we conduct several ablation experiments. The parametres to be explored include the size of the input image, the number of skip-connections, and the patch size.

\begin{table*}[htbp]
	\centering 
	\caption{\label{tab:test}Comparison of performance of different input image sizes. }
	\scalebox{1}[1]{
		\resizebox{\textwidth}{!}{
			\begin{tabular}{lcccccccccc} 
				\toprule 
				Resolution & DSC  & Aorta & Gallbladder & Kidney(L) & Kidney(R) & Liver & Pancreas & Spleen & Stomach  \\ 
				\midrule 
				$224 \times 224$  & 78.09  & 85.87 & 61.38 & 84.83 & 79.36 & 94.28 & 57.65 & 87.74 & 73.55  \\ 
				$512 \times 512$ & 80.39  & 90.00 & 56.86 & 83.27 & 76.21 & 95.06 & 67.76 & 91.16 & 82.82 \\ 
				
				\bottomrule 
	\end{tabular}}}
	
\end{table*}

\begin{table*}[htbp]
	\centering 
	\caption{\label{tab:test}Comparison of performance of different patch sizes.}
	\scalebox{1}[1]{
		\resizebox{\textwidth}{!}{
			\begin{tabular}{lcccccccccc} 
				\toprule 
				Patch size  & DSC  & Aorta & Gallbladder & Kidney(L) & Kidney(R) & Liver & Pancreas & Spleen & Stomach  \\ 
				\midrule 
				$16 \times 16$  & 78.09  & 85.87 & 61.38 & 84.83 & 79.36 & 94.28 & 57.65 & 87.74 & 73.55  \\ 
				$20 \times 20$ & 77.92  & 85.96 & 61.81 & 84.66 & 76.67 & 94.28 & 55.94 & 86.90 & 77.13 \\ 
				$24 \times 24$ & 77.27  & 86.84 & 60.41 & 84.05 & 77.83 & 93.74 & 56.16 & 86.33 & 72.77 \\
				
				\bottomrule 
	\end{tabular}}}
	
\end{table*}

1) Effect of the input size: when the image size is larger, it often contains more pixel information. We compared model performance  when the input image size is $224\times224$ and $512\times512$. Due to the limitation of computing resources, the batch size is set to $24$ and $6$. It can be seen from Table 2 that when the image resolution increases, the DSC will increase significantly. It may be that as the image size increases, the patch size is unchanged, the length of the sequence increases, so that more global and detailed information in the image can be learned and converted into more effective encoding results, thereby improving segmentation performance. However, the increasing of image size needs more computing resources, longer training time. Taking into account the calculation cost and other factors, the input image size we finally adopted is $224 \times 224$.

2) Effect of the number of skip-connections: by setting the number of skip-connections to $0$, $1$, $2$, and $3$, we can explore its impact on model performance. Fig. 3 shows the DSC index results of 8 organs segmentation under different numbers of skip-connections. It can be seen that as the number of skip-connections decreases, the effect of segmentation becomes worse. The skip-connections can effectively combine the encoding part, the bottom upsampling part and the decoding part, so that the model is more robust and can achieve better performance. It is worth noting that when the skip-connections reduce to $0$, for Arota and Pancreas, the DSC drops by $10.8\%$ and $10.16\%$ respectively.

3) Effect of the patch size and sequence length: the length of the transformer encoding sequence depends on the size of the patch, and the sequence length has a certain impact on the acquisition of image features. The use of different patch sizes will have a greater impact on segmentation performance. According to the existing knowledge, the product of the square of the patch size and the length of the transformer sequence is a fixed value. When the patch size is larger, the sequence is shorter, and the effective information that can be learned between encoders is less, which affects the acquisition of global spatial features. When the patch size is smaller, as the sequence grows, the parameters that need to be learned grow rapidly, which greatly occupies computing resources. Therefore, it is important to choose the suitable patch size. In this experiment, we use $16\times 16$ patches. It can be seen from Table 3 that when the patch size is $16$, the DSC is $78.09\%$, and when the patch size is $24$, the DSC is $77.27\%$. The growth rate of DSC is limited, but the occupied memory shows a multiple growth.

\section{CONCLUSION}
\label{sec:foot}
In this paper, we propose a TransClaw U-Net network structure, which combines the convolution operation of CNN and the transformer mechanism, and has achieved excellent performance on the Synapse multi-organ CT dataset. Compared with the traditional CNN-network, the multi-head self-attention in the transformer can effectively obtain the global context information and make up for the short-comings of the convolution operation in this respect. In the encoding part, these two parts are combined with each other to achieve better results. The decoding part retains the  upsampling part, which makes the network segmentation of medical images more accurate. Ablation experimental results show that TransClaw U-Net has excellent generalization performance. 

\begin{figure*}[htbp] 
\centering 
	
	\includegraphics[height=8cm, width=18cm]{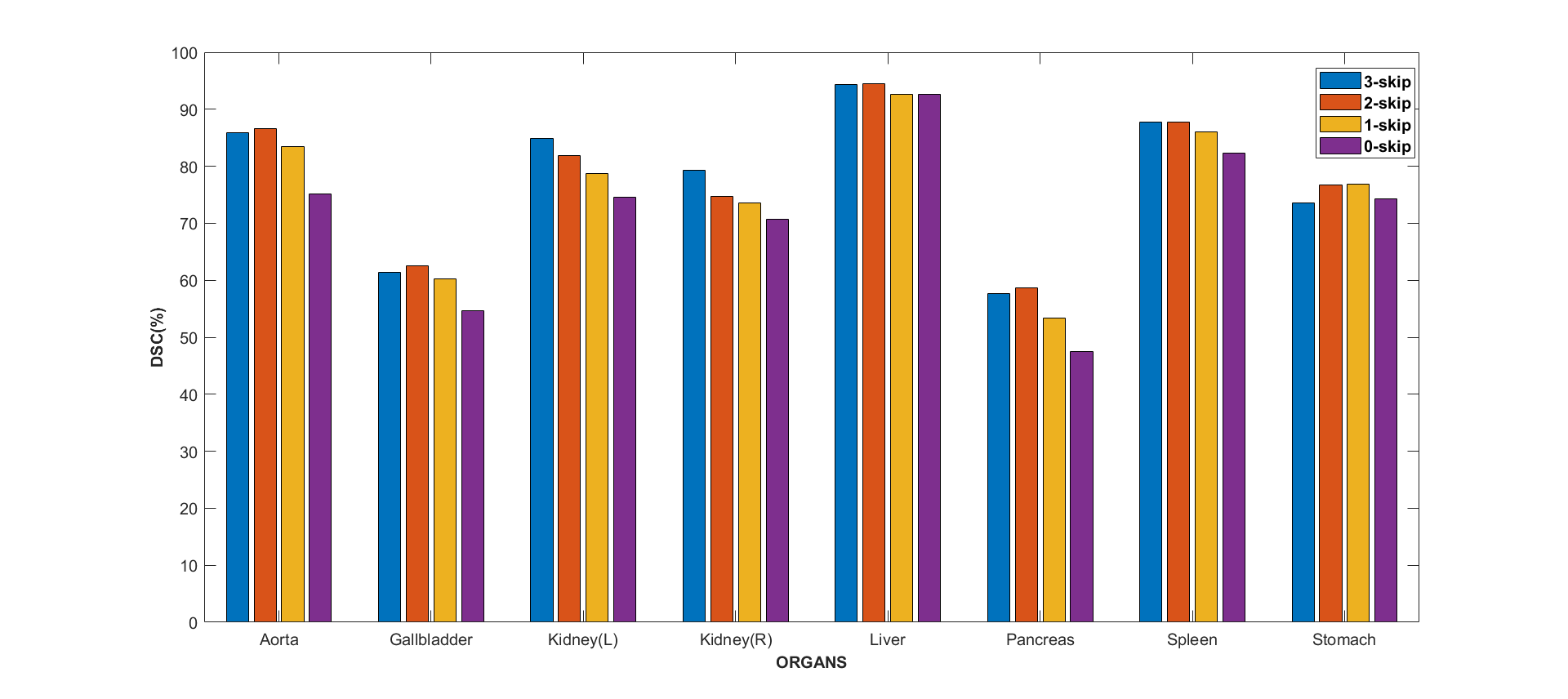} 
	
	\caption{Comparison of performance of different numbers of skip-connections.} \label{fig:graph} \end{figure*}

\vfill
\pagebreak

\bibliographystyle{IEEEbib}
\bibliography{strings,refs}

\begin{thebibliography}{10}

\bibitem{long2015fully}
Jonathan Long, Evan Shelhamer, and Trevor Darrell,
\newblock ``Fully convolutional networks for semantic segmentation,''
\newblock in {\em Proceedings of the IEEE conference on computer vision and
  pattern recognition}, 2015, pp. 3431--3440.

\bibitem{hatamizadeh2021unetr}
Ali Hatamizadeh, Dong Yang, Holger Roth, and Daguang Xu,
\newblock ``Unetr: Transformers for 3d medical image segmentation,''
\newblock {\em arXiv preprint arXiv:2103.10504}, 2021.

\bibitem{ronneberger2015u}
Olaf Ronneberger, Philipp Fischer, and Thomas Brox,
\newblock ``U-net: Convolutional networks for biomedical image segmentation,''
\newblock in {\em International Conference on Medical image computing and
  computer-assisted intervention}. Springer, 2015, pp. 234--241.

\bibitem{yao2020claw}
Chang Yao, Jingyu Tang, Menghan Hu, Yue Wu, Wenyi Guo, Qingli Li, and Xiao-Ping
  Zhang,
\newblock ``Claw u-net: A unet-based network with deep feature concatenation
  for scleral blood vessel segmentation,''
\newblock {\em arXiv preprint arXiv:2010.10163}, 2020 (This paper has been
  accepted in 2021 CICAI.).

\bibitem{zhou2019unet++}
Zongwei Zhou, Md~Mahfuzur~Rahman Siddiquee, Nima Tajbakhsh, and Jianming Liang,
\newblock ``Unet++: Redesigning skip connections to exploit multiscale features
  in image segmentation,''
\newblock {\em IEEE transactions on medical imaging}, vol. 39, no. 6, pp.
  1856--1867, 2019.

\bibitem{xiao2018weighted}
Xiao Xiao, Shen Lian, Zhiming Luo, and Shaozi Li,
\newblock ``Weighted res-unet for high-quality retina vessel segmentation,''
\newblock in {\em 2018 9th international conference on information technology
  in medicine and education (ITME)}. IEEE, 2018, pp. 327--331.

\bibitem{chen2019channel}
Yilong Chen, Kai Wang, Xiangyun Liao, Yinling Qian, Qiong Wang, Zhiyong Yuan,
  and Pheng-Ann Heng,
\newblock ``Channel-unet: a spatial channel-wise convolutional neural network
  for liver and tumors segmentation,''
\newblock {\em Frontiers in genetics}, vol. 10, pp. 1110, 2019.

\bibitem{oktay2018attention}
Ozan Oktay, Jo~Schlemper, Loic~Le Folgoc, Matthew Lee, Mattias Heinrich,
  Kazunari Misawa, Kensaku Mori, Steven McDonagh, Nils~Y Hammerla, Bernhard
  Kainz, et~al.,
\newblock ``Attention u-net: Learning where to look for the pancreas,''
\newblock {\em arXiv preprint arXiv:1804.03999}, 2018.

\bibitem{alom2018recurrent}
Md~Zahangir Alom, Mahmudul Hasan, Chris Yakopcic, Tarek~M Taha, and Vijayan~K
  Asari,
\newblock ``Recurrent residual convolutional neural network based on u-net
  (r2u-net) for medical image segmentation,''
\newblock {\em arXiv preprint arXiv:1802.06955}, 2018.

\bibitem{li2018h}
Xiaomeng Li, Hao Chen, Xiaojuan Qi, Qi~Dou, Chi-Wing Fu, and Pheng-Ann Heng,
\newblock ``H-denseunet: hybrid densely connected unet for liver and tumor
  segmentation from ct volumes,''
\newblock {\em IEEE transactions on medical imaging}, vol. 37, no. 12, pp.
  2663--2674, 2018.

\bibitem{zhou2017fixed}
Yuyin Zhou, Lingxi Xie, Wei Shen, Yan Wang, Elliot~K Fishman, and Alan~L
  Yuille,
\newblock ``A fixed-point model for pancreas segmentation in abdominal ct
  scans,''
\newblock in {\em International conference on medical image computing and
  computer-assisted intervention}. Springer, 2017, pp. 693--701.

\bibitem{yu2017automatic}
Lequan Yu, Jie-Zhi Cheng, Qi~Dou, Xin Yang, Hao Chen, Jing Qin, and Pheng-Ann
  Heng,
\newblock ``Automatic 3d cardiovascular mr segmentation with densely-connected
  volumetric convnets,''
\newblock in {\em International conference on medical image computing and
  computer-assisted intervention}. Springer, 2017, pp. 287--295.

\bibitem{chen2021transunet}
Jieneng Chen, Yongyi Lu, Qihang Yu, Xiangde Luo, Ehsan Adeli, Yan Wang, Le~Lu,
  Alan~L Yuille, and Yuyin Zhou,
\newblock ``Transunet: Transformers make strong encoders for medical image
  segmentation,''
\newblock {\em arXiv preprint arXiv:2102.04306}, 2021.

\bibitem{schlemper2019attention}
Jo~Schlemper, Ozan Oktay, Michiel Schaap, Mattias Heinrich, Bernhard Kainz, Ben
  Glocker, and Daniel Rueckert,
\newblock ``Attention gated networks: Learning to leverage salient regions in
  medical images,''
\newblock {\em Medical image analysis}, vol. 53, pp. 197--207, 2019.

\bibitem{gu2019net}
Zaiwang Gu, Jun Cheng, Huazhu Fu, Kang Zhou, Huaying Hao, Yitian Zhao, Tianyang
  Zhang, Shenghua Gao, and Jiang Liu,
\newblock ``Ce-net: Context encoder network for 2d medical image
  segmentation,''
\newblock {\em IEEE transactions on medical imaging}, vol. 38, no. 10, pp.
  2281--2292, 2019.

\bibitem{zhao2017pyramid}
Hengshuang Zhao, Jianping Shi, Xiaojuan Qi, Xiaogang Wang, and Jiaya Jia,
\newblock ``Pyramid scene parsing network,''
\newblock in {\em Proceedings of the IEEE conference on computer vision and
  pattern recognition}, 2017, pp. 2881--2890.

\bibitem{devlin2018bert}
Jacob Devlin, Ming-Wei Chang, Kenton Lee, and Kristina Toutanova,
\newblock ``Bert: Pre-training of deep bidirectional transformers for language
  understanding,''
\newblock {\em arXiv preprint arXiv:1810.04805}, 2018.

\bibitem{radford2018improving}
Alec Radford, Karthik Narasimhan, Tim Salimans, and Ilya Sutskever,
\newblock ``Improving language understanding by generative pre-training,''
\newblock 2018.

\bibitem{liu2021swin}
Ze~Liu, Yutong Lin, Yue Cao, Han Hu, Yixuan Wei, Zheng Zhang, Stephen Lin, and
  Baining Guo,
\newblock ``Swin transformer: Hierarchical vision transformer using shifted
  windows,''
\newblock {\em arXiv preprint arXiv:2103.14030}, 2021.

\bibitem{vaswani2017attention}
Ashish Vaswani, Noam Shazeer, Niki Parmar, Jakob Uszkoreit, Llion Jones,
  Aidan~N Gomez, Lukasz Kaiser, and Illia Polosukhin,
\newblock ``Attention is all you need,''
\newblock {\em arXiv preprint arXiv:1706.03762}, 2017.

\bibitem{dosovitskiy2020image}
Alexey Dosovitskiy, Lucas Beyer, Alexander Kolesnikov, Dirk Weissenborn,
  Xiaohua Zhai, Thomas Unterthiner, Mostafa Dehghani, Matthias Minderer, Georg
  Heigold, Sylvain Gelly, et~al.,
\newblock ``An image is worth 16x16 words: Transformers for image recognition
  at scale,''
\newblock {\em arXiv preprint arXiv:2010.11929}, 2020.

\bibitem{cao2021swin}
Hu~Cao, Yueyue Wang, Joy Chen, Dongsheng Jiang, Xiaopeng Zhang, Qi~Tian, and
  Manning Wang,
\newblock ``Swin-unet: Unet-like pure transformer for medical image
  segmentation,''
\newblock {\em arXiv preprint arXiv:2105.05537}, 2021.

\bibitem{tsai2003shape}
Andy Tsai, Anthony Yezzi, William Wells, Clare Tempany, Dewey Tucker, Ayres
  Fan, W~Eric Grimson, and Alan Willsky,
\newblock ``A shape-based approach to the segmentation of medical imagery using
  level sets,''
\newblock {\em IEEE transactions on medical imaging}, vol. 22, no. 2, pp.
  137--154, 2003.

\bibitem{xie2017aggregated}
Saining Xie, Ross Girshick, Piotr Doll{\'a}r, Zhuowen Tu, and Kaiming He,
\newblock ``Aggregated residual transformations for deep neural networks,''
\newblock in {\em Proceedings of the IEEE conference on computer vision and
  pattern recognition}, 2017, pp. 1492--1500.

\bibitem{zhu2019deformable}
Xizhou Zhu, Han Hu, Stephen Lin, and Jifeng Dai,
\newblock ``Deformable convnets v2: More deformable, better results,''
\newblock in {\em Proceedings of the IEEE conference on computer vision and
  pattern recognition}, 2019, pp. 9308--9316.

\bibitem{hu2018relation}
Han Hu, Jiayuan Gu, Zheng Zhang, Jifeng Dai, and Yichen Wei,
\newblock ``Relation networks for object detection,''
\newblock in {\em Proceedings of the IEEE Conference on Computer Vision and
  Pattern Recognition}, 2018, pp. 3588--3597.

\bibitem{wang2018non}
Xiaolong Wang, Ross Girshick, Abhinav Gupta, and Kaiming He,
\newblock ``Non-local neural networks,''
\newblock in {\em Proceedings of the IEEE conference on computer vision and
  pattern recognition}, 2018, pp. 7794--7803.

\bibitem{touvron2020training}
Hugo Touvron, Matthieu Cord, Matthijs Douze, Francisco Massa, Alexandre
  Sablayrolles, and Herv{\'e} J{\'e}gou,
\newblock ``Training data-efficient image transformers \& distillation through
  attention,''
\newblock {\em arXiv preprint arXiv:2012.12877}, 2020.

\bibitem{milletari2016v}
Fausto Milletari, Nassir Navab, and Seyed-Ahmad Ahmadi,
\newblock ``V-net: Fully convolutional neural networks for volumetric medical
  image segmentation,''
\newblock in {\em 2016 fourth international conference on 3D vision (3DV)}.
  IEEE, 2016, pp. 565--571.

\bibitem{fu2020domain}
Shuhao Fu, Yongyi Lu, Yan Wang, Yuyin Zhou, Wei Shen, Elliot Fishman, and Alan
  Yuille,
\newblock ``Domain adaptive relational reasoning for 3d multi-organ
  segmentation,''
\newblock in {\em International Conference on Medical Image Computing and
  Computer-Assisted Intervention}. Springer, 2020, pp. 656--666.

\end{thebibliography}

\end{document}